\documentclass[10pt,twocolumn,letterpaper]{article}

\usepackage{cvpr}
\usepackage{times}
\usepackage{epsfig}
\usepackage{graphicx}
\usepackage{amsmath}
\usepackage{amssymb}

\usepackage{stfloats}
\usepackage[x11names,dvipsnames,table]{xcolor} %for use in color links
\usepackage{multirow}
\usepackage{array}
\newcolumntype{L}[1]{>{\raggedright\let\newline\\\arraybackslash\hspace{0pt}}m{#1}}
\newcolumntype{C}[1]{>{\centering\let\newline\\\arraybackslash\hspace{0pt}}m{#1}}
\newcolumntype{R}[1]{>{\raggedleft\let\newline\\\arraybackslash\hspace{0pt}}m{#1}}

\usepackage[breaklinks=true,bookmarks=false]{hyperref}

\cvprfinalcopy % *** Uncomment this line for the final submission

\def\cvprPaperID{****} % *** Enter the CVPR Paper ID here

\begin{document}

%%%%%%%%% TITLE
\title{Training Deep Neural Networks in Generations:\\A More Tolerant Teacher Educates Better Students}

\author{Chenglin Yang, Lingxi Xie, Siyuan Qiao, Alan Yuille\\
The Johns Hopkins University\\
{\tt\small \{chenglin.yangw,198808xc,alan.l.yuille\}@gmail.com}\quad{\tt\small siyuan.qiao@jhu.edu}
}

\maketitle
%\thispagestyle{empty}

%%%%%%%%% ABSTRACT
\begin{abstract}
We focus on the problem of training a deep neural network {\bf in generations}. The flowchart is that, in order to optimize the target network (student), another network (teacher) with the same architecture is first trained, and used to provide part of supervision signals in the next stage. While this strategy leads to a higher accuracy, many aspects (e.g., why teacher-student optimization helps) still need further explorations.

This paper studies this problem from a perspective of controlling the strictness in training the teacher network. Existing approaches mostly used a hard distribution (e.g., one-hot vectors) in training, leading to a strict teacher which itself has a high accuracy, but we argue that the teacher needs to be more tolerant, although this often implies a lower accuracy. The implementation is very easy, with merely an extra loss term added to the teacher network, facilitating a few secondary classes to emerge and complement to the primary class. Consequently, the teacher provides a milder supervision signal (a less peaked distribution), and makes it possible for the student to learn from inter-class similarity and potentially lower the risk of over-fitting. Experiments are performed on standard image classification tasks (CIFAR100 and ILSVRC2012). Although the teacher network behaves less powerful, the students show a persistent ability growth and eventually achieve higher classification accuracies than other competitors. Model ensemble and transfer feature extraction also verify the effectiveness of our approach.
\end{abstract}

%%%%%%%%% BODY TEXT
\section{Introduction}
\label{Introduction}

\vspace{0.2cm}
\noindent
$\bullet$\quad{\em ``Indigo comes from blue, but it is bluer than blue.''}
\vspace{-0.2cm}
\begin{flushright}
---\textsc{An Old Proverb}
\end{flushright}

Deep learning, especially the convolutional neural networks, has been widely applied to computer vision problems. Among them, image classification has been considered the fundamental task which sets the backbones vision systems for other problems~\cite{krizhevsky2012imagenet}\cite{simonyan2015very}\cite{szegedy2015going}\cite{he2016deep}, and the knowledge or features extracted from these modules are transferrable for generic image representation purposes~\cite{razavian2014cnn} or other vision tasks~\cite{long2015fully}\cite{ren2015faster}\cite{xie2015holistically}\cite{newell2016stacked}.

A fundamental task in computer vision is to optimize deep networks for image classification. Most existing work achieved this goal by fitting the outputs of a model to {\em one-hot vectors}. For each training sample $\left(\mathbf{x}_n,y_n\right)$ where $\mathbf{x}_n$ is an image matrix and $y_n$ is the class label (out of $C$ classes), the goal is to find network parameters $\boldsymbol{\theta}$, so that ${\mathbf{f}\!\left(\mathbf{x}_n;\boldsymbol{\theta}\right)}\approx{\left[0,\ldots,1,\ldots,0\right]^\top}\in{\mathbb{R}^C}$, {\em i.e.}, only the $y_n$-th dimension is $1$ and all others are $0$. Despite its effectiveness, this is not necessarily the optimal target to fit, because except for maximizing the confidence score of the {\em primary} class ({\em i.e.}, the ground-truth), allowing for some {\em secondary} classes ({\em i.e.}, those visually similar ones to the ground-truth) to be preserved may help to alleviate the risk of over-fitting. Some proposed to learn a class-level similarity matrix~\cite{deng2010what}\cite{verma2012learning}\cite{wu2017hierarchical}, but these approaches are unable to capture inter-class similarities at the image level, {\em e.g.}, different {\em cat} images may be visually similar to different classes.

We turn to an alternative solution, {\em i.e.}, extracting knowledge from a trained ({\em teacher}) network and guide another ({\em student}) network in an individual training process. This algorithm, known as {\em knowledge distillation}, was first used to network compression (the student network is much smaller than the teacher network, but can achieve a comparable accuracy)~\cite{hinton2015distilling}, but later it was verified effective in the scenario that teacher and student networks have the same architecture, in which the student is expected to achieve a higher accuracy than the teacher~\cite{furlanello2018born}. Despite its effectiveness, it remains unclear how teacher-student optimization works, and if we can find some key factors to guide the design of such optimization processes.

This paper provides an interesting perspective, focusing on the {\em strictness} of the teacher. We argue that classification accuracy is not the major goal of the teacher network; instead, it is designed to be tolerant ({\em i.e.}, producing less peaked distributions of confidence) so that the students can learn inter-class similarity and potentially prevent over-fitting. To achieve this goal, we add an extra term to the standard cross-entropy loss in training the teacher network, facilitating it to distribute confidence to a few secondary classes. Although this harms the accuracy of the teacher network, it indeed provides more room for the student network(s), and eventually, the students are better than those educated by a strict teacher. In standard image classification experiments on CIFAR100 and ILSVRC2012, our approach reports higher classification accuracy than its competitors, regardless using a single model, an ensemble of multiple models, or transferring trained models for feature extraction.

The contribution of this work is three-fold. First, we propose a new perspective to interpret why teacher-student optimization works. Second, we suggest an evaluation method to quantize its impact. Third, we design an efficient ``tolerant-teacher'' framework which achieves superior performance.

The reminder of this paper is organized as follows. We first review related work in the next section, and then illustrate our approach. After experiments are shown, we conclude our work in the final section.

\section{Related Work}
\label{RelatedWork}

\subsection{Deep Learning and Neural Networks}
\label{RelatedWork:DeepNetworks}

Deep learning has been dominating the field of computer vision. Powered by large-scale image datasets~\cite{deng2009imagenet} and powerful computational resources, it is possible to train very deep networks for various computer vision tasks. The fundamental idea of deep learning is to design a hierarchical structure containing multiple {\em layers}, each of which contains a number of {\em neurons} having the same or similar mathematical functions. Researchers believe that a sufficiently deep network is able to fit very complicated distributions in the feature space. In a fundamental task known as image classification, deep neural networks~\cite{krizhevsky2012imagenet} have achieved much higher accuracy than conventional handcrafted features~\cite{perronnin2010improving}. Towards better recognition performance, researchers designed deeper and deeper networks~\cite{simonyan2015very}\cite{szegedy2015going}\cite{he2016deep}\cite{huang2017densely}\cite{hu2018squeeze}, and even proposed to automatically explore network architectures~\cite{xie2017genetic}\cite{zoph2017neural}.

The rapid progress of deep learning has helped a lot of computer vision tasks. Features extracted from trained classification networks can be transferred to small datasets for image classification~\cite{donahue2014decaf}, retrieval~\cite{razavian2014cnn} or object detection~\cite{girshick2014rich}. An even more effective way is to insert specified network modules for these tasks, and initializing these models with part of the weights learned for image classification. This flowchart, often referred to as fine-tuning, works well in a variety of problems, including object detection\cite{girshick2015fast}\cite{ren2015faster}, semantic segmentation~\cite{long2015fully}\cite{chen2016deeplab}, edge detection~\cite{xie2015holistically}, {\em etc}.

\subsection{Training Very Deep Networks in Generations}
\label{RelatedWork:InGenerations}

Deep network optimization has become an important yet challenging problem. Training very deep neural networks ({\em e.g.}, more than $100$ layers) requires specifically designed techniques to assist numerical stability, such as ReLU activation~\cite{nair2010rectified}, Dropout~\cite{srivastava2014dropout} and batch normalization~\cite{ioffe2015batch}. However, as depth increases, the large number of parameters makes neural networks easy to be over-fitted, especially when the amount of training data is limited. Therefore, it is often instructive to introduce extra {\em priors} to constrain the training process and thus prevent over-fitting. A common prior assumes that some classes are visually or semantically similar~\cite{deng2010what}, and adds a class-level similarity matrix to the loss function~\cite{verma2012learning}\cite{wu2017hierarchical}, but it is unable to deal with per-image similarity which is well noted in previous research~\cite{wang2014learning}\cite{akata2016label}\cite{zhang2018image}.

An effective way to solve this issue is {\em teacher-student optimization}, in which a teacher network is trained beforehand, and then used to guide the optimization of a student network. Thus, the output of the teacher network carries class-level similarity in its confidence score for each image. Previously, teacher-student optimization was used to distill knowledge from a larger network and then compress it into a smaller network~\cite{hinton2015distilling}, or initialize a deeper network with pre-trained weights of a shallower network~\cite{romero2014fitnets}\cite{chen2015net2net}\cite{simonyan2015very}. This basic idea was then extended in many ways, including using various ways of supervision~\cite{szegedy2016rethinking}\cite{pereyra2017regularizing}, using multiple teachers to provide a better guidance~\cite{tarvainen2017mean}, adding supervision in intermediate neural responses~\cite{yim2017gift}, and allowing two networks to help optimize each other~\cite{zhang2017deep}. In a recent work named the born-again network~\cite{furlanello2018born}, this method was used to optimize the same network {\em in generations}, in which the next generation was guided by two terms, namely, the standard cross-entropy loss and the KL-divergence between the teacher and student signals.

\section{Our Approach}
\label{Approach}

This section presents our approach. We first introduce a framework of optimizing neural networks {\em in generations}, and then provide an empirical analysis on the benefits of such optimization methods, and suggest a quantitative way of evaluating this process. Based on these, we finally design a flowchart, which train a tolerant teacher network to improve the overall performance of optimization.

\subsection{Teacher-Student Optimization}
\label{Approach:Motivation}

We consider a standard network optimization task. Given a model $\mathbb{M}$ which has a parameterized form of ${\mathbf{y}}={\mathbf{f}\!\left(\mathbf{x};\boldsymbol{\theta}\right)}$, where $\mathbf{x}$ and $\mathbf{y}$ are input and output, and $\boldsymbol{\theta}$ denotes learnable parameters ({\em e.g.}, convolutional weights). Given a training set ${\mathcal{D}}={\left\{\left(\mathbf{x}_1,\mathbf{y}_1\right),\ldots,\left(\mathbf{x}_N,\mathbf{y}_N\right)\right\}}$, the goal is to determine the parameter $\boldsymbol{\theta}$ that best fits these data.

One of the most popular optimization methods starts with setting all weights as random noise $\boldsymbol{\theta}^{\left(0\right)}$, and then applies gradient descent to update them gradually. Each time, a subset $\mathcal{B}$ is sampled from $\mathcal{D}$, and a loss function computed according to the difference between prediction and labels:
\begin{equation}
\label{Eqn:StandardLoss}
{\mathcal{L}\!\left(\mathcal{B};\boldsymbol{\theta}\right)}={-\frac{1}{\left|\mathcal{B}\right|}{\sum_{\left(\mathbf{x}_n,\mathbf{y}_n\right)\in\mathcal{B}}}\mathbf{y}_n^\top\ln\mathbf{f}\!\left(\mathbf{x}_n;\boldsymbol{\theta}\right)}.
\end{equation}
We can interpret this process as a heuristic way of searching over the high-dimensional parameter space defined by the network $\mathbf{f}\!\left(\mathbf{x};\boldsymbol{\theta}\right)$. However, due to the complicated network design and limited dataset size, this training process often suffers over-fitting, {\em i.e.}, a $\boldsymbol{\theta}$ is found to achieve a high accuracy on the training set, but the testing accuracy is still far below the training accuracy. This limits us from generalizing the trained model to unobserved testing data.

One way of softening supervision label signals is to perform {\em teacher-student optimization}. In this process, a teacher model $\mathbb{M}^\mathrm{T}$: $\mathbf{f}\!\left(\mathbf{x};\boldsymbol{\theta}^\mathrm{T}\right)$ is first trained in the same dataset using Eqn~\eqref{Eqn:StandardLoss}, and then used to train a student model $\mathbb{M}^\mathrm{S}$: $\mathbf{f}\!\left(\mathbf{x};\boldsymbol{\theta}^\mathrm{S}\right)$ using a mixture loss~\cite{furlanello2018born}:
\begin{eqnarray}
\label{Eqn:StudentLoss}
\nonumber
{\mathcal{L}^\mathrm{S}\!\left(\mathcal{B};\boldsymbol{\theta}^\mathrm{S}\right)}={-\frac{1}{\left|\mathcal{B}\right|}{\sum_{\left(\mathbf{x}_n,\mathbf{y}_n\right)\in\mathcal{B}}}\left\{\lambda\cdot\mathbf{y}_n^\top\ln\mathbf{f}\!\left(\mathbf{x}_n;\boldsymbol{\theta}^\mathrm{S}\right)+\right.}\\
\quad{\left.\left(1-\lambda\right)\cdot\mathrm{KL}\!\left[\mathbf{f}\!\left(\mathbf{x}_n;\boldsymbol{\theta}^\mathrm{T}\right)\|\mathbf{f}\!\left(\mathbf{x}_n;\boldsymbol{\theta}^\mathrm{S}\right)\right]\right\}}.
\end{eqnarray}
This is to say, the teacher network provides $\mathbf{f}\!\left(\mathbf{x}_n;\boldsymbol{\theta}^\mathrm{T}\right)$, a softened version of the one-hot vector $\mathbf{y}_n$, so that the student network can find a compromise between these two signals. In the next subsection, we will show how this formulation helps in training a better student.

A straightforward extension of teacher-student optimization is to allow a network to be optimized {\em in generations}. This requires training a {\em patriarch} model, denoted by $\mathbb{M}^{\left(0\right)}$, which is only supervised by the dataset. $M$ more generations follow, in which the $m$-th generation trains a student $\mathbb{M}^{\left(m\right)}$ with the supervision of a teacher $\mathbb{M}^{\left(m-1\right)}$. Most often~\cite{furlanello2018born}, the recognition accuracy goes up in the first few generations, but starts to saturate and go down. We will analyze the reason in the following parts.

\subsection{Preserving Secondary Information: An Important Factor in Teacher-Student Optimization}
\label{Approach:Analysis}

Previously, teacher-student optimization was mostly applied to distill knowledge from a larger network, so that it can be compressed into a smaller network with recognition accuracy largely preserved~\cite{hinton2015distilling}, or applied to initializing a deeper network with pre-trained weights from a shallower network~\cite{romero2014fitnets}\cite{chen2015net2net}. As the first work to train an identical network in generations, \cite{furlanello2018born} explained the benefit as a weighted balance between the ground-truth (one-hot) signal and the teacher signal, but it did not notice an important role of the teacher: suggesting class-level similarity.

\newcommand{\colwidthA}{0.90cm}
\begin{table}[!btp]
\centering{
\setlength{\tabcolsep}{0.12cm}
\begin{tabular}{|l||R{\colwidthA}|R{\colwidthA}|R{\colwidthA}|R{\colwidthA}||R{\colwidthA}|R{\colwidthA}|}
\hline
{}       & Top-$1$ & Top-$2$ & Top-$3$ & Top-$4$ & Train & Test    \\
\hline\hline
Gen \#0  & $99.28$ & $0.57$ & $0.09$ & $0.03$ & $99.74$ & $71.55$ \\
\hline
Gen \#1  & $98.68$ & $1.00$ & $0.18$ & $0.06$ & $99.63$ & $71.41$ \\
\hline
Gen \#2  & $98.42$ & $1.13$ & $0.23$ & $0.09$ & $99.60$ & $72.30$ \\
\hline
Gen \#3  & $98.33$ & $1.19$ & $0.24$ & $0.09$ & $99.62$ & $72.26$ \\
\hline
Gen \#4  & $98.28$ & $1.24$ & $0.25$ & $0.09$ & $99.59$ & $72.52$ \\
\hline
\end{tabular}}
\caption{Confidence distribution ($\%$) on top-$4$ classes (individually determined for each training sample), obtained in a born-again process (with one patriarch and $5$ more generations), training a $110$-layer ResNet on CIFAR100. We also show training and testing accuracies ($\%$) in the last columns.}
\label{Tab:TopConfidenceScores}
\end{table}

\newcommand{\widthA}{3.5cm}
\begin{figure}[!t]
\centering
\includegraphics[width=\widthA]{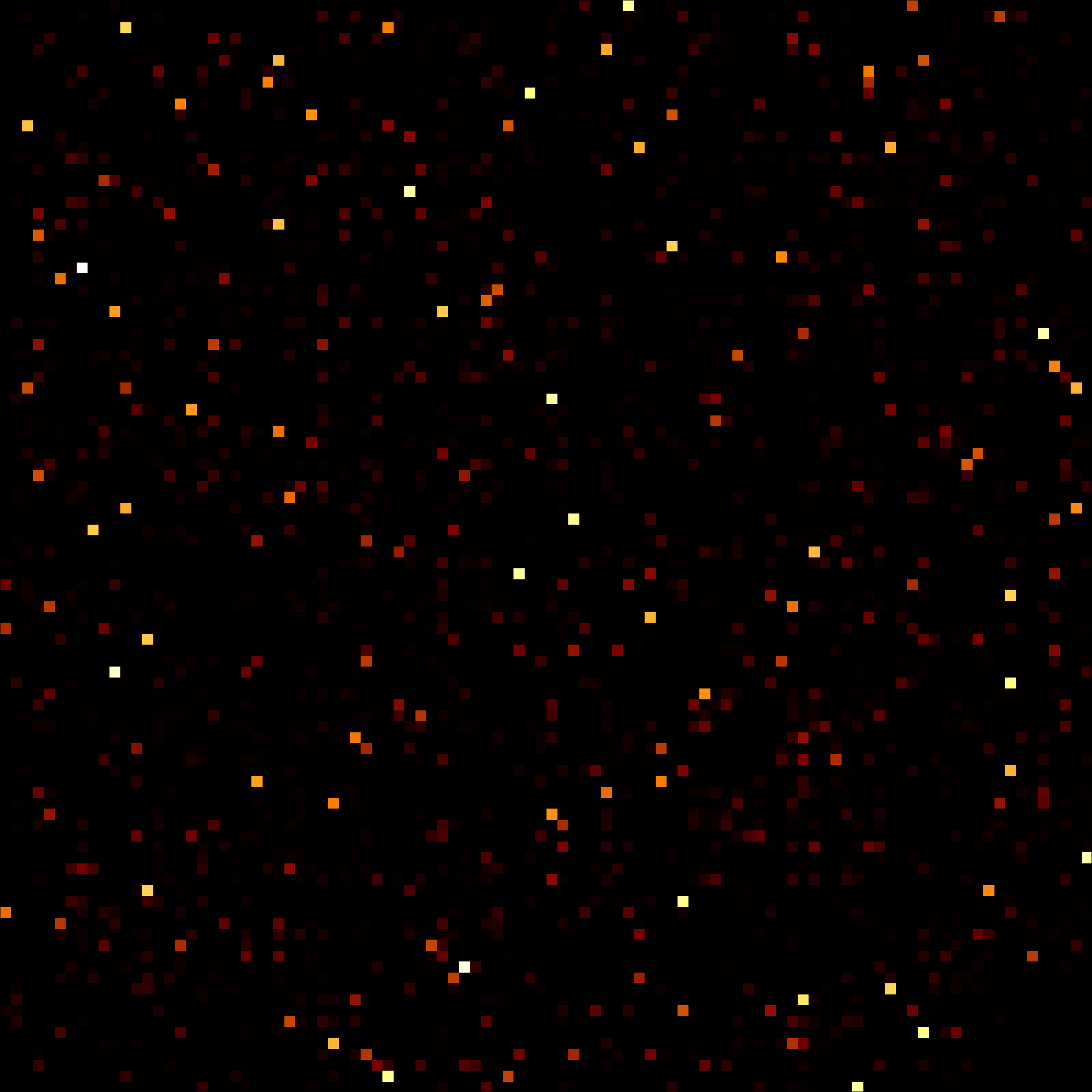}\hspace{0.5cm}
\includegraphics[width=\widthA]{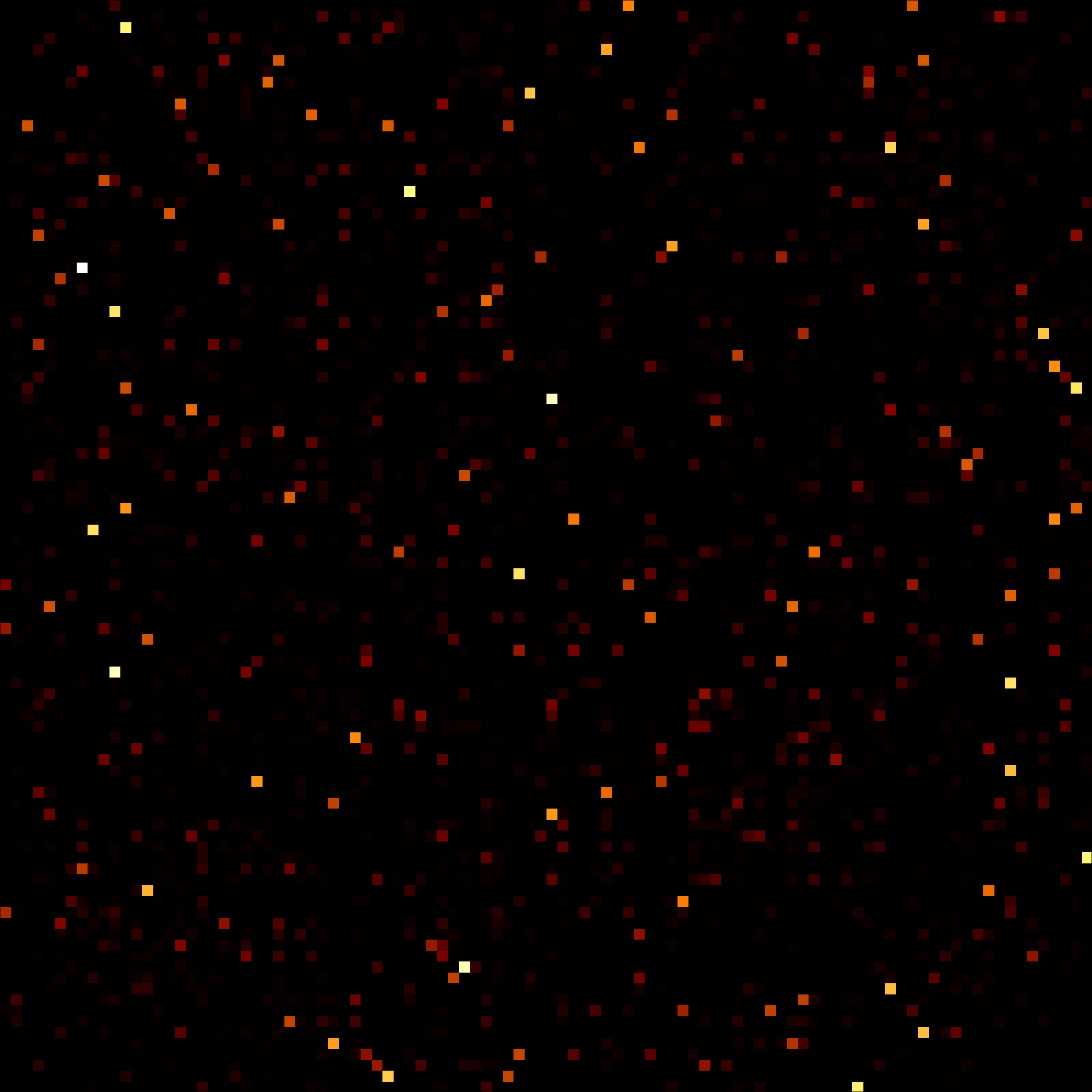}\\
\vspace{0.5cm}
\includegraphics[width=\widthA]{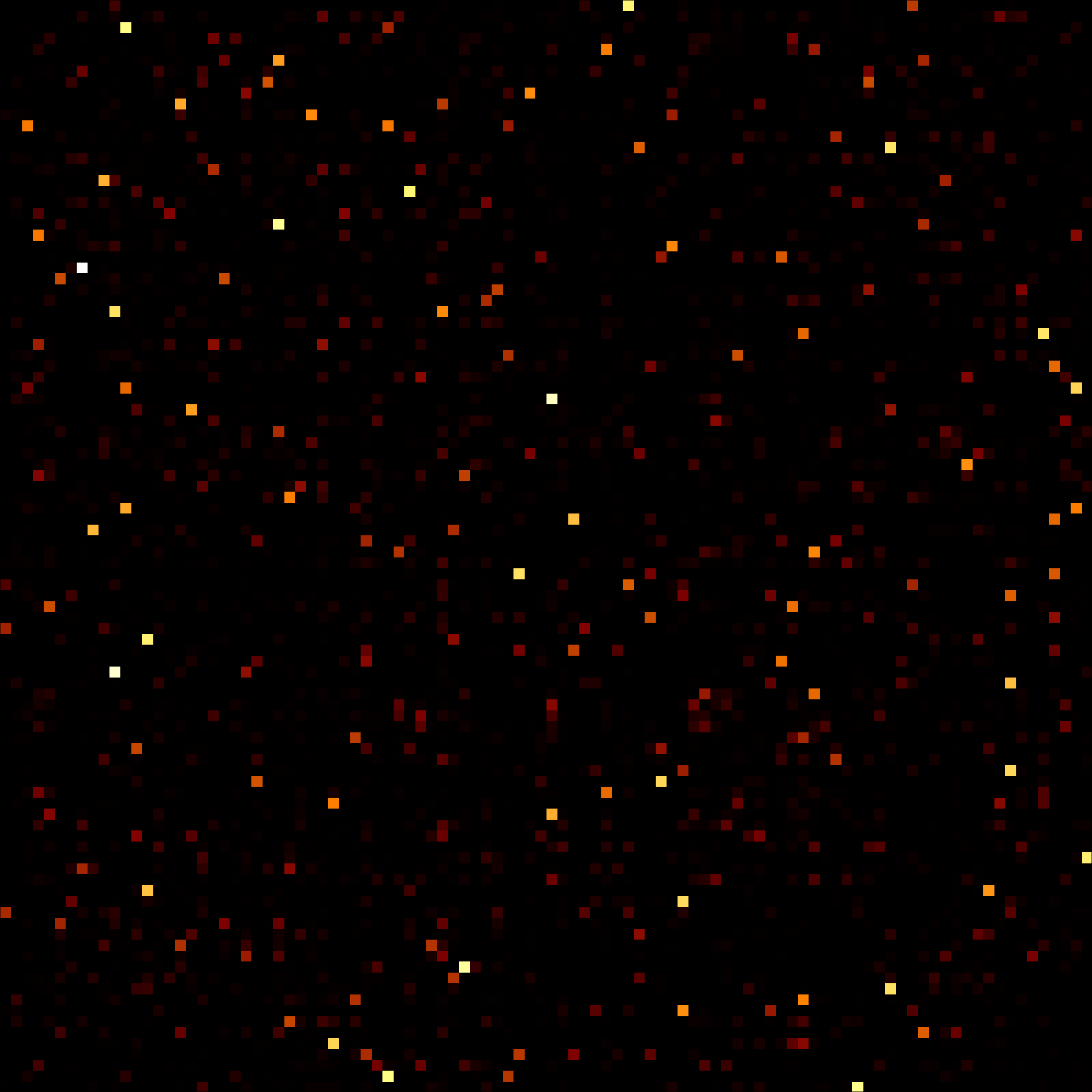}\hspace{0.5cm}
\includegraphics[width=\widthA]{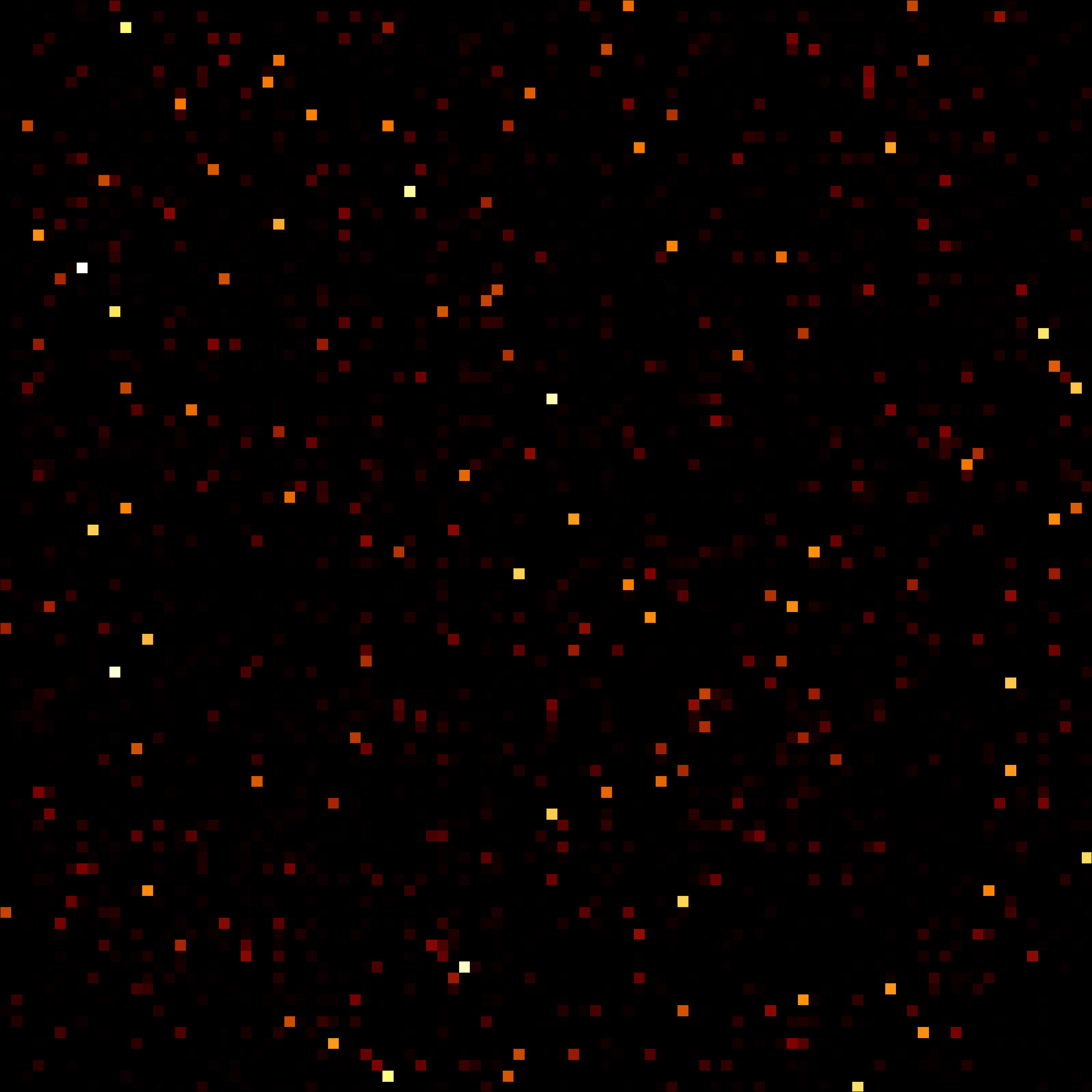}\\
\vspace{0.5cm}
\includegraphics[width=\widthA]{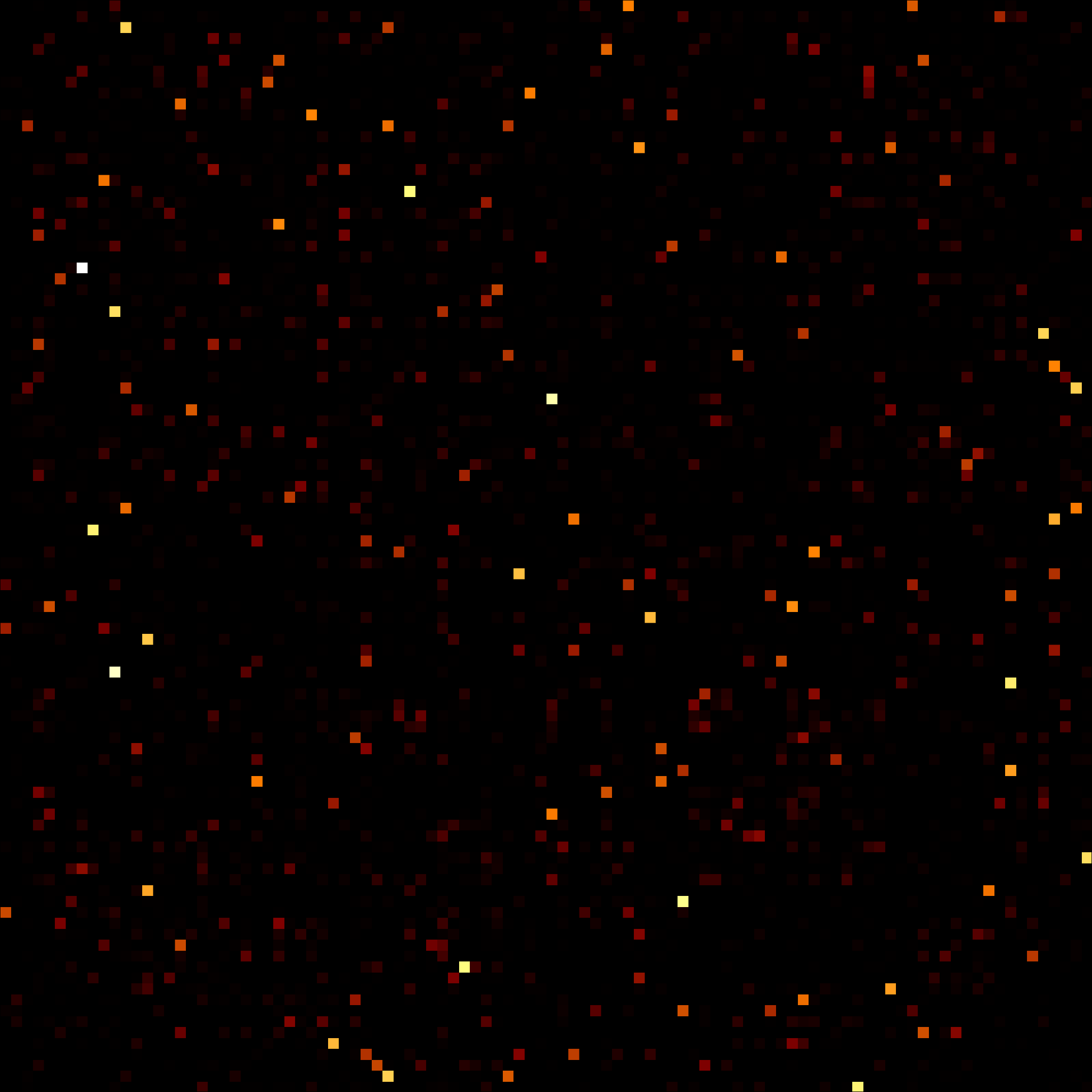}\hspace{0.5cm}
\includegraphics[width=\widthA]{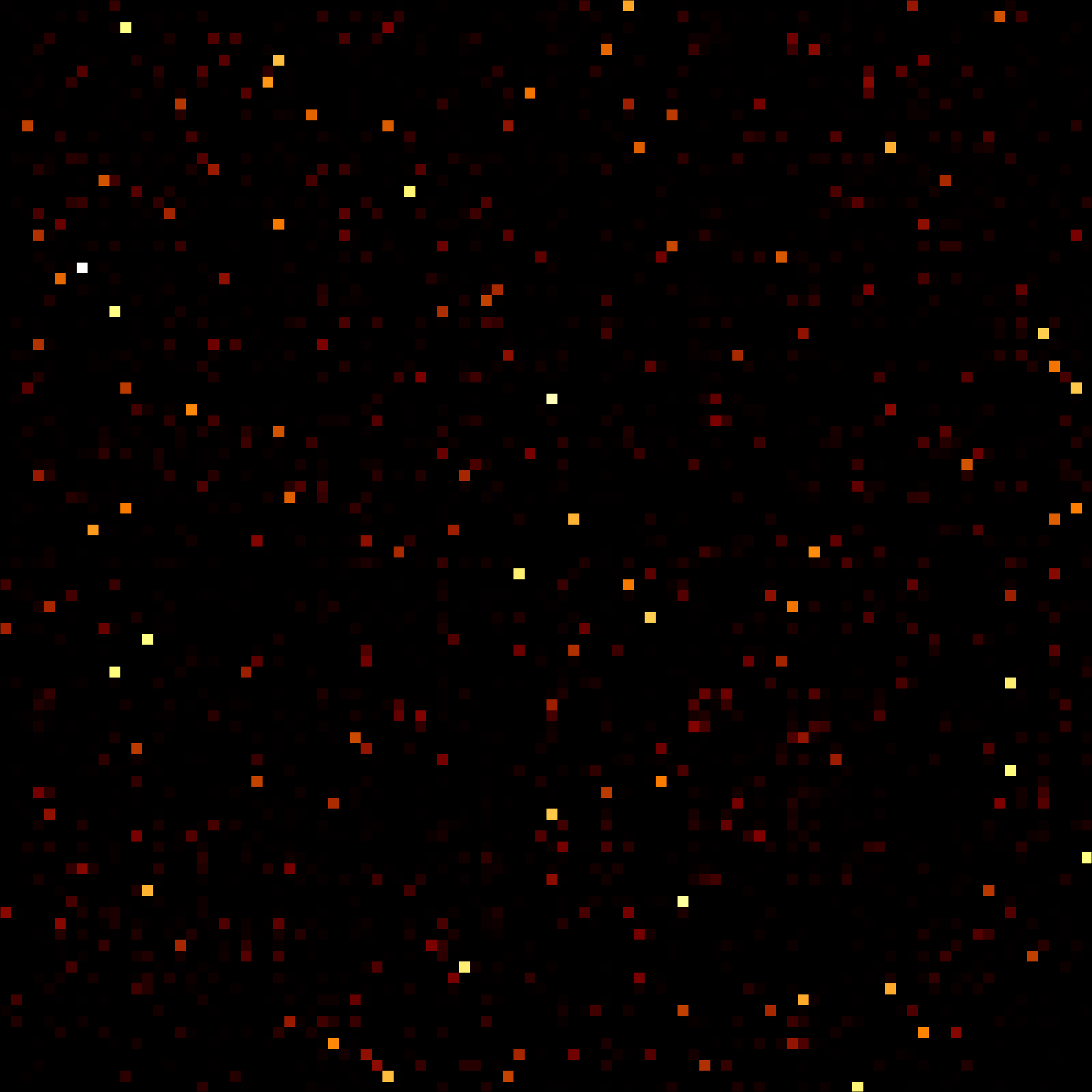}
\caption{The $100\times100$ confusion matrices produced by the patriarch and the first five students of a born-again process, training a $110$-layer ResNet on CIFAR100. See Table~\ref{Tab:TopConfidenceScores} for quantitative numbers. The rows in these matrices indicate the ground-truth class, and the columns indicate the class with the second highest confidence score. The color of a cell is closer to yellow when the corresponding value is larger.}
\label{Fig:ConfusionMatrices}
\end{figure}

To reveal this, we investigate network training of a born-again process~\cite{furlanello2018born}, with a $110$-layer ResNet optimized on CIFAR100. In Table~\ref{Tab:TopConfidenceScores}, we list the training and testing accuracies in each generation. Guided by softened distributions, the students achieve higher recognition performance than the patriarch.

We hence ask a question: what is the key benefit of being trained by a softened label distribution? To answer it, we perform statistics on the class with second highest confidence score, and plot the results as a confusion matrix in Figure~\ref{Fig:ConfusionMatrices}. We find that a deep network is able to automatically learn semantically similar classes {\em for each image individually}\footnote{For example, {\em cat} images are often considered similar to {\em dog}, but sometimes {\em deer} becomes the most similar class; {\em automobile} is most similar to {\em truck} in $60\%$ of time, but in another $19\%$ and $7\%$ of time, it is most similar to {\em ship} and {\em airplane}, respectively.}. We name it as {\bf secondary information}, which corresponds to the primary information provided by supervision. By taking these image-dependent information, the student network can avoid being fit to unnecessarily strict distributions and thus generalizes better. This motivates us to design a mechanism to measure the quality of secondary information and then try to construct better teacher signals.

\subsection{Towards High-Quality Secondary Information}
\label{Approach:SecondaryInformation}

First, we note that the key to finding secondary information is to soften the output feature vector. We investigate three ways to achieve this goal. The first two methods follows two pieces of prior work named label smoothing regularization (LSR)~\cite{szegedy2016rethinking} and confidence penalty (CP)~\cite{pereyra2017regularizing}. Both of them added an extra term to the original cross-entropy loss, pushing the score distribution (after softmax) towards less peaked at the primary class. In LSR, the added term is the KL-divergence between the score distribution and the uniform distribution, while in CP, it is the negative entropy gain. These two methods have a common drawback: they facilitate the confidence scores to distribute over all classes, regardless if these classes are visually similar to the training sample.

As the third option, we propose a more reasonable approach. Instead of computing an extra loss over all classes, we pick up a few classes which have been assigned with the highest confidence scores, and assume that these classes are more likely to be semantically similar to the input image. We set a fixed integer $K$ which stands for the number of semantically reasonable classes for each image, including the primary class\footnote{Using a fixed $K$ may not be optimal, but it simplifies our approach and also works sufficiently well in either network training or transferring a trained network to other recognition tasks (see experiments).}. Then, we compute the gap between the confidence scores of the primary class and other $K-1$ classes with highest scores:
\begin{eqnarray}
\label{Eqn:PatriarchLoss}
\nonumber
{\mathcal{L}^\mathrm{T}\!\left(\mathcal{B};\boldsymbol{\theta}^\mathrm{T}\right)}={\frac{1}{\left|\mathcal{B}\right|}{\sum_{\left(\mathbf{x}_n,\mathbf{y}_n\right)\in\mathcal{B}}}\left\{-\eta\cdot\mathbf{y}_n^\top\ln\mathbf{f}\!\left(\mathbf{x}_n;\boldsymbol{\theta}^\mathrm{T}\right)+\right.}\\
\quad{\left.\left(1-\eta\right)\cdot\left[f_{a_1}-\frac{1}{K-1}{\sum_{k=2}^K}f_{a_k}\right]\right\}}.
\end{eqnarray}
We name this method as top score difference (TSD), where $f_{a_k}^\mathrm{T}$ is short for the $k$-th largest element of $\mathbf{f}\!\left(\mathbf{x}_n;\boldsymbol{\theta}^\mathrm{T}\right)$, and $\eta$ is a hyper-parameter controlling the balance between the ground-truth supervision and the score penalty term.

We evaluate these three options, as well as the baseline (using one-hot vectors, Eqn~\eqref{Eqn:StandardLoss}), on their ability of preserving secondary information. To this end, we train a $110$-layer ResNet on CIFAR100. Detailed settings can be found in the {\bf Experiments} section. We perform four individual training-in-generation processes, with the only difference lying in the patriarchs, {\em i.e.}, four different options are applied. All the remaining generations follow Eqn~\eqref{Eqn:StudentLoss} with ${\lambda}={0.6}$. To maximally make fair comparison, we guarantee the same initialization weights for each model.

According to our conjecture, a good patriarch should preserve more secondary information, and thus is less discriminative in fine-level classes. We measure this factor at the class level. After these networks are trained, we analyze their behavior on both training and testing sets. For each image, the neural responses from the last residual block ($8\times8\times64$) are average-pooled to obtain a $64$-dimensional vector. The training and testing sets of CIFAR100 has $50\rm{,}000$ and $10\rm{,}000$ images, which are uniformly distributed over $100$ classes. These $100$ classes are partitioned into $20$ superclasses, with $5$ fine-level classes in each superclass $\mathcal{S}_j$. We compute the mean vector for each class, denoted by $\mathbf{v}_i^\mathrm{C}$, ${i}={1,2,\ldots,100}$, and for each superclass, denoted by $\mathbf{v}_j^\mathrm{S}$, ${j}={1,2,\ldots,20}$, respectively.

Based on these, we compute two statistics within each superclass and among different superclasses, respectively. The first one measures how much feature vector $\mathbf{v}_i^\mathrm{C}$ differs from the mean of its superclass, $\mathbf{v}_j^\mathrm{S}$ where ${i}\in{\mathcal{S}_j}$. Similarly, the second one measures the difference between each $\mathbf{v}_j^\mathrm{S}$ and the overall average vector ${\mathbf{v}^\mathrm{A}}={\frac{1}{20}{\sum_j}\mathbf{v}_j^\mathrm{S}}$. Mathematically, these two metrics are:
\begin{eqnarray}
\label{Eqn:SimilarityMetrics}
\nonumber
{\mathrm{Dist}^\mathrm{C}} & = & {\frac{1}{100}\times{\sum_{i=1}^{100}}\arccos\frac{\left\langle\mathbf{v}_i^\mathrm{C},\mathbf{v}_{j\left(i\right)}^\mathrm{S}\right\rangle}{\left\|\mathbf{v}_i^\mathrm{C}\right\|\cdot\left\|\mathbf{v}_{j\left(i\right)}^\mathrm{S}\right\|}},\\
{\mathrm{Dist}^\mathrm{S}} & = & {\frac{1}{20}\times{\sum_{j=1}^{20}}\arccos\frac{\left\langle\mathbf{v}_j^\mathrm{S},\mathbf{v}^\mathrm{A}\right\rangle}{\left\|\mathbf{v}_j^\mathrm{S}\right\|\cdot\left\|\mathbf{v}^\mathrm{A}\right\|}},
\end{eqnarray}
where $j\!\left(i\right)$ denotes the superclass index for class $i$.

\renewcommand{\colwidthA}{1.15cm}
\newcommand{\colwidthB}{1.30cm}
\begin{table}[!btp]
\centering{
\setlength{\tabcolsep}{0.08cm}
\begin{tabular}{|l||R{\colwidthA}|R{\colwidthA}||R{\colwidthA}|R{\colwidthA}|R{\colwidthB}|}
\hline
{}        & \multicolumn{2}{c||}{Training}                          & \multicolumn{3}{c|}{Testing}                                                 \\
\cline{2-6}
{}        & $\mathrm{Dist}^\mathrm{C}$ & $\mathrm{Dist}^\mathrm{S}$ & $\mathrm{Dist}^\mathrm{C}$ & $\mathrm{Dist}^\mathrm{S}$ & Best Acc           \\
\hline\hline
BL        & $0.3813$                   & $0.4202$                   & $0.3234$                   & $0.3881$                   & $72.61\%$          \\
\hline\hline
LSR       & $0.6182$                   & $0.5475$                   & $0.4875$                   & $0.4805$                   & $73.46\%$          \\
\hline
CP        & $0.3829$                   & $0.4114$                   & $0.3267$                   & $0.3815$                   & $72.86\%$          \\
\hline\hline
TSD-$0.6$ & $0.3461$                   & $\mathbf{0.5759}$          & $0.3153$                   & $\mathbf{0.5349}$          & $\mathbf{73.72\%}$ \\
\hline
TSD-$0.7$ & $\mathbf{0.3433}$          & $0.5274$                   & $\mathbf{0.3086}$          & $0.4894$                   & $73.18\%$          \\
\hline
TSD-$0.8$ & $0.3499$                   & $0.4782$                   & $0.3097$                   & $0.4449$                   & $73.39\%$          \\
\hline
\end{tabular}}
\caption{Statistics on different patriarchs. The definitions of $\mathrm{Dist}^\mathrm{C}$ and $\mathrm{Dist}^\mathrm{S}$ are provided in Eqn~\eqref{Eqn:SimilarityMetrics}. BL is for baseline (as in~\cite{furlanello2018born}), LSR for label smoothing regularization, CP for confidence penalty, and TSD-$u\!\left(\eta\right)$ for top-score difference with a parameter $u\!\left(\eta\right)$. We also report the best testing accuracy throughout $10$ generations, which reflect the potential of the patriarch model.}
\label{Tab:DistanceMetrics}
\end{table}

Results are summarized in Table~\ref{Tab:DistanceMetrics}. Compared to the baseline, LSR features are much more discriminative (both $\mathrm{Dist}^\mathrm{C}$ and $\mathrm{Dist}^\mathrm{S}$ are much larger), while CP does not behave much differently. TSD-$0.6$ ($u\!\left(\eta\right)$ is an intuitive way of parameterizing $\eta$ -- see the next part for details) best satisfies our assumption: increasing $\mathrm{Dist}^\mathrm{S}$ so that coarse-level classification becomes better, meanwhile decreasing $\mathrm{Dist}^\mathrm{C}$ so that reasonable secondary information is preserved and learned by students. Using TSD-$0.7$ or TSD-$0.8$ does not heavily impact $\mathrm{Dist}^\mathrm{C}$, but causes $\mathrm{Dist}^\mathrm{S}$ to be much smaller. The best classification accuracy is obtained by TSD-$0.6$.

Based on these observations, we can conclude that: {\bf in teacher-student optimization, the student learns best from a teacher that preserves reasonable secondary information.} Hence, we suggest a framework that starts with a tolerant patriarch to optimize deep networks in generations.

\renewcommand{\widthA}{5.8cm}
\begin{figure*}[!t]
\centering
\includegraphics[width=\widthA]{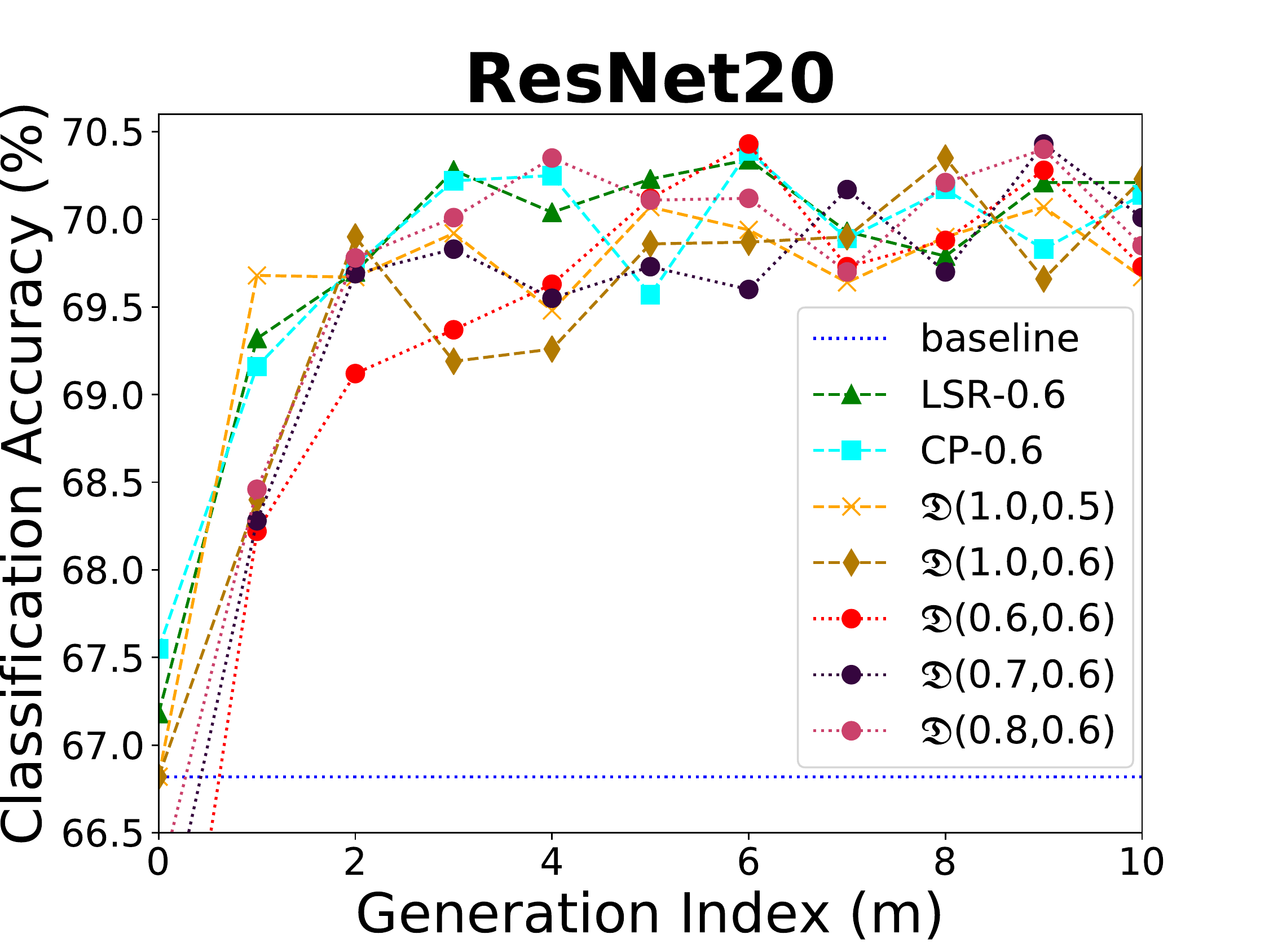}\hfill
\includegraphics[width=\widthA]{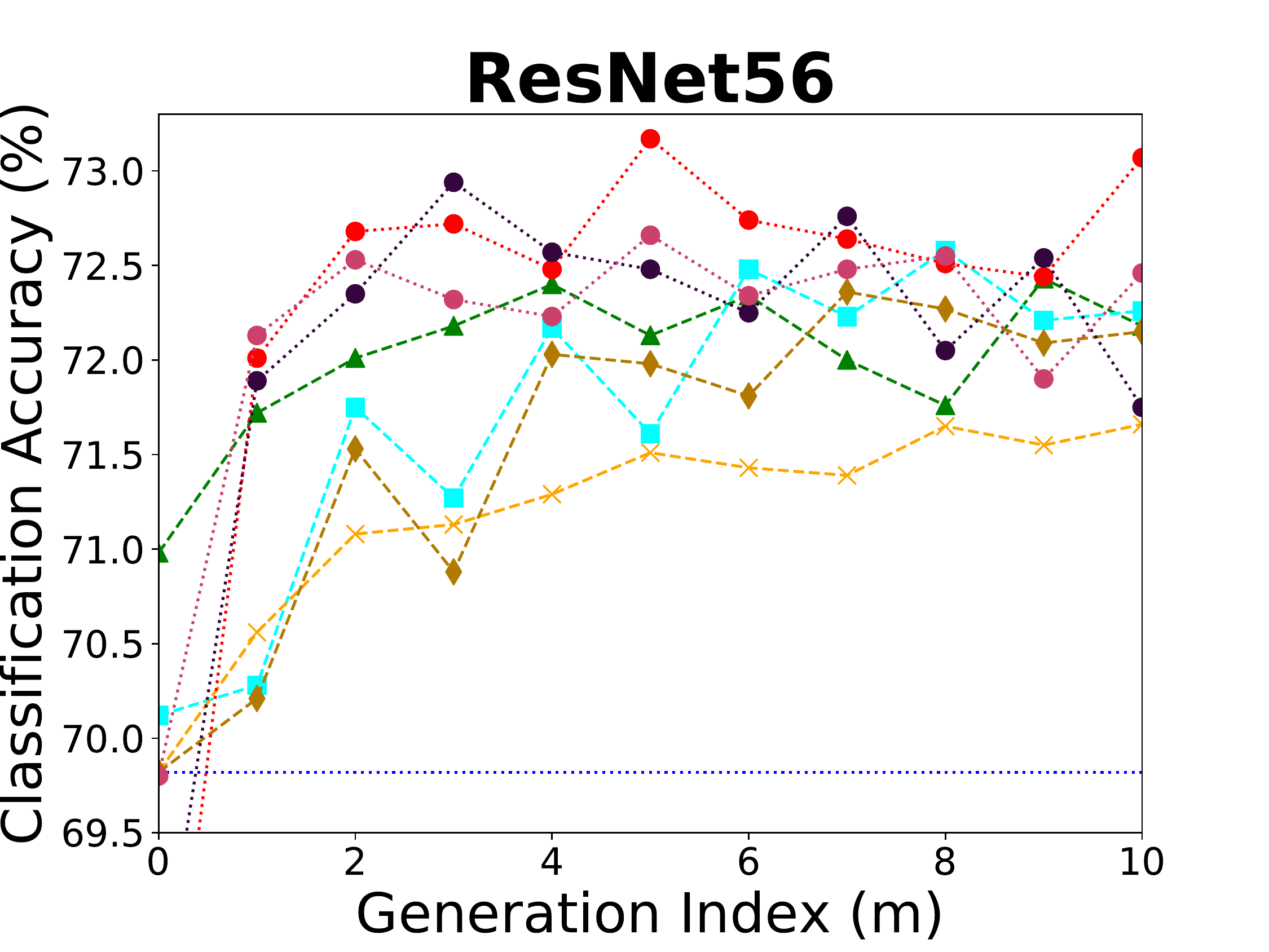}\hfill
\includegraphics[width=\widthA]{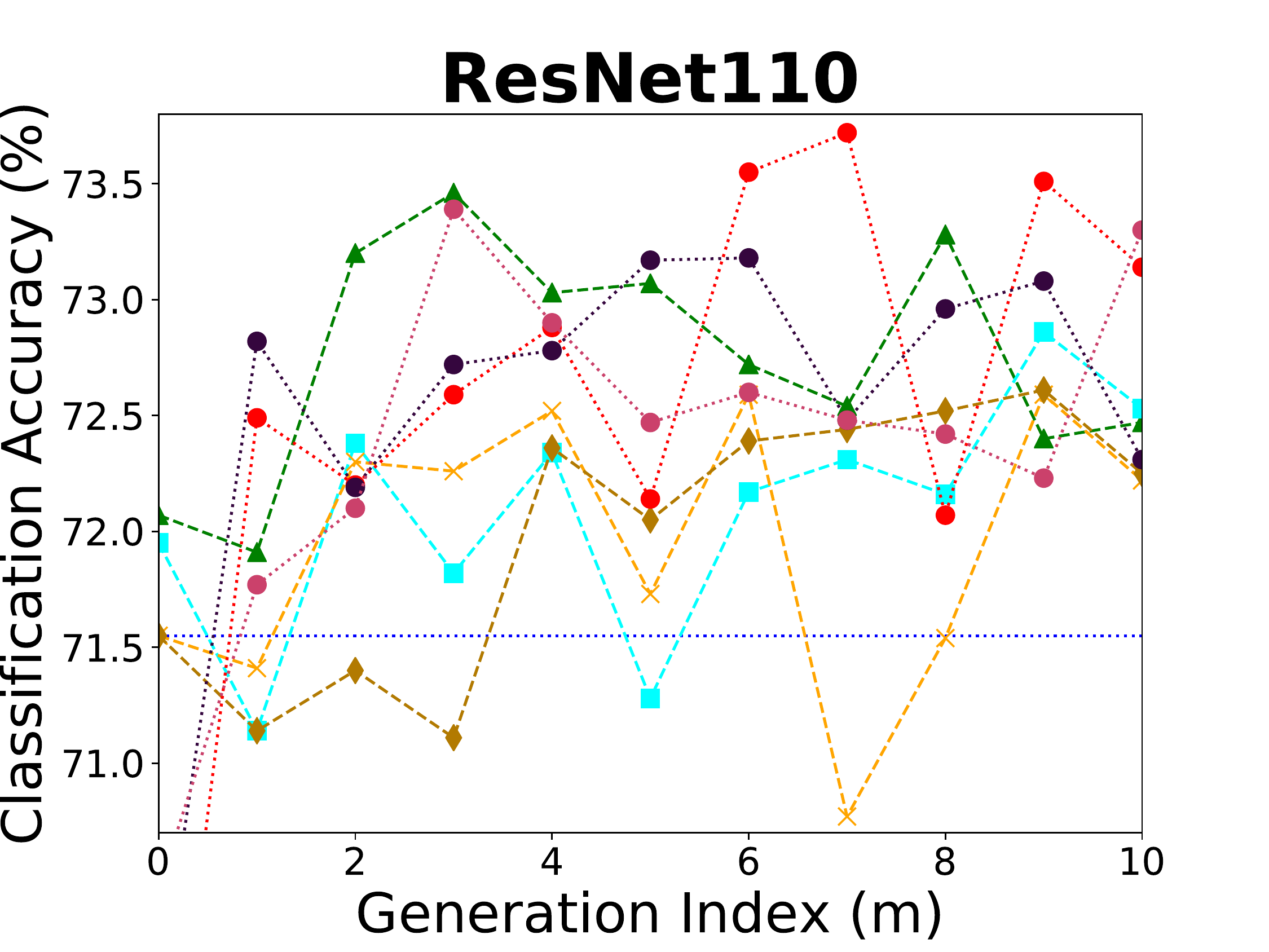}
\caption{Classification accuracy ($\%$) on CIFAR100, produced by different training-in-generation processes. The baseline approach (single generation) corresponds to $\mathfrak{D}\!\left(1.0,0.0\right)$, and $\mathfrak{D}\!\left(1.0,0.5\right)$ and $\mathfrak{D}\!\left(1.0,0.6\right)$ are born-again networks. LSR-$0.6$ and CP-$0.6$ indicate replacing the patriarch model with label smoothing regularization and confidence penalty, and use ${\lambda}={0.6}$ in generations. All three plots share the same legend (shown in the first plot).}
\label{Fig:CIFAR100ResNets}
\end{figure*}

\subsection{Details of Training in Generations}
\label{Approach:Details}

We set Eqn~\eqref{Eqn:PatriarchLoss} to be the loss function to train the patriarch model $\mathbb{M}^{\left(0\right)}$. Mathematically, to minimize Eqn~\eqref{Eqn:PatriarchLoss}, $\mathbf{f}\!\left(\mathbf{x}_n;\boldsymbol{\theta}^\mathrm{T}\right)$ shall satisfy ${0}<{f_{a_1}^\mathrm{T}}<{1}$, ${f_{a_2}^\mathrm{T}+\ldots+f_{a_K}^\mathrm{T}}={1-f_{a_1}^\mathrm{T}}$, and all other entries are $0$. We can derive the optimal ${f_{a_1}^\mathrm{T}}\doteq{u\!\left(\eta\right)}={\min\left\{\frac{\eta}{1-\eta}\cdot\frac{K-1}{K},1\right\}}$, which is a monotonically increasing function with respect to $\eta$. Therefore, it is equivalent to consider $u\!\left(\eta\right)$ instead of $\eta$. In experiments, we shall see that a deeper network often requires a larger $u\!\left(\eta\right)$, or equivalently a larger $\eta$, in order to get better trained.

Two side notes are made on Eqn~\eqref{Eqn:PatriarchLoss} and the hyper-parameter $K$. First, our formulation does not guarantee the primary class $a_1$ corresponds to the true class. But as we shall see in experiments, after a sufficient number of training epochs, the training accuracy is always close to $100\%$. Second, $K$ is often difficult to estimate, and may vary among different primary classes. In practice, we fix ${K}={5}$ for simplicity\footnote{This fits CIFAR100 well, because it contains $20$ coarse groups and each of them has $5$ finer-level classes. This setting also works well in ILSVRC2012, although there are different numbers of semantically similar classes for each class. We transfer the trained models on ILSVRC2012 to other recognition tasks to reveal the generalization properties. See experiments for details.}.

Then, $M$ generations follow the patriarch model. At the $m$-th generation, $\mathbb{M}^{\left(m\right)}$ learns from $\mathbb{M}^{\left(m-1\right)}$ using Eqn~\eqref{Eqn:StudentLoss}, with models $\mathbf{f}\!\left(\mathbf{x}_n;\boldsymbol{\theta}^\mathrm{T}\right)$ and $\mathbf{f}\!\left(\mathbf{x}_n;\boldsymbol{\theta}^\mathrm{S}\right)$ replaced by $\mathbf{f}\!\left(\mathbf{x}_n;\boldsymbol{\theta}^{\left(m-1\right)}\right)$ and $\mathbf{f}\!\left(\mathbf{x}_n;\boldsymbol{\theta}^{\left(m\right)}\right)$, respectively. Similarly, the optimal $\mathbf{f}\!\left(\mathbf{x}_n;\boldsymbol{\theta}^{\left(m\right)}\right)$ to minimize $\mathcal{L}^{\left(m\right)}$ shall satisfy ${0}<{f_{a_1}^{\left(m\right)}}<{1}$, ${f_{a_2}^{\left(m\right)}+\ldots+f_{a_K}^{\left(m\right)}}={\left(1-f_{a_1}^{\left(m\right)}\right)}$, and all other entries are $0$. So, we have ${f_{a_1}^{\left(m\right)}}\doteq{w\!\left(\lambda,\mathbf{f}^{\left(m-1\right)}\right)}$ which is monotonically increasing with respect to $\lambda$.

In summary, the entire optimization process is parameterized by $K$, $\eta$ and $\lambda$. We fix ${K}={5}$ and use $u\!\left(\eta\right)$ to equivalently replace $\eta$, so that each process is denoted by $\mathfrak{D}\!\left(u\!\left(\eta\right),\lambda\right)$. As special cases of our approach, the conventional network optimization process can be abbreviated as $\mathfrak{D}\!\left(1.0,0.0\right)$, and training a born-again network~\cite{furlanello2018born} corresponds to $\mathfrak{D}\!\left(1.0,0.5\right)$.

Last but not least, given that the trained model $\mathbf{f}\!\left(\cdot;\boldsymbol{\theta}\right)$ has a sufficient ability of fitting data, we have ${f_{a_1}^{\left(m\right)}}\rightarrow{1}$ as ${m}\rightarrow{+\infty}$, regardless of $u\!\left(\eta\right)$ and $\lambda$. Therefore, the secondary information in teacher signals is gradually weakened, and the effect of optimizing Eqn~\eqref{Eqn:StudentLoss} is more and more similar to optimizing Eqn~\eqref{Eqn:StandardLoss}. This implies that recognition accuracy may saturate and start to descend after a few generations (see Figure~\ref{Fig:CIFAR100ResNets} for experimental results).

\section{Experiments}
\label{Experiments}

\subsection{The CIFAR100 Dataset}
\label{Experiments:CIFAR100}

\renewcommand{\colwidthA}{2.10cm}
\renewcommand{\colwidthB}{3.80cm}
\newcommand{\colwidthC}{1.00cm}
\begin{table*}[!btp]
\centering{
\setlength{\tabcolsep}{0.12cm}
\begin{tabular}{|l||R{\colwidthA}|R{\colwidthA}|R{\colwidthA}|R{\colwidthA}|R{\colwidthA}|R{\colwidthA}|}
\hline
                                          &
\multicolumn{1}{c|}{Gen \#0} & \multicolumn{1}{c|}{Gen \#1} & \multicolumn{1}{c|}{Gen \#2} &
\multicolumn{1}{c|}{Gen \#3} & \multicolumn{1}{c|}{Gen \#4} & \multicolumn{1}{c|}{Gen \#5} \\
\hline\hline
Baseline ($100$ layers)                   &
         $22.20$ ($22.89$) &                        $-$ &                        $-$ &                        $-$ &                        $-$ &                        $-$ \\
\hline
\quad$\mathfrak{D}\!\left(0.6,0.6\right)$ &
         $23.96$ ($25.00$) &          $21.29$ ($21.34$) & $\mathbf{20.51}$ ($21.59$) &          $20.83$ ($20.99$) &          $21.01$ ($21.53$) &          $21.27$ ($21.61$) \\
\hline
\quad\quad$+$Ensemble                     &
                       $-$ &                    $20.20$ &                    $18.38$ &                    $17.79$ &                    $17.37$ &           $\mathbf{17.25}$ \\
\hline
\quad$\mathfrak{D}\!\left(0.7,0.6\right)$ &
         $22.98$ ($23.43$) &          $21.24$ ($21.50$) &          $21.48$ ($21.80$) & $\mathbf{20.94}$ ($21.47$) &          $21.51$ ($21.69$) &          $21.87$ ($22.28$) \\
\hline
\quad\quad$+$Ensemble                     &
                       $-$ &                    $19.63$ &                    $18.83$ &                    $17.70$ &                    $17.56$ &           $\mathbf{17.23}$ \\
\hline\hline
Baseline ($190$ layers)                   &
         $17.22$ ($17.62$) &                        $-$ &                        $-$ &                        $-$ &                        $-$ &                        $-$ \\
\hline
\quad$\mathfrak{D}\!\left(0.6,0.6\right)$ &
         $18.87$ ($19.40$) &          $17.42$ ($17.99$) &          $17.26$ ($18.00$) &          $17.13$ ($17.52$) &          $17.24$ ($17.75$) & $\mathbf{17.01}$ ($17.22$) \\
\hline
\quad\quad$+$Ensemble                     &
                       $-$ &                    $16.83$ &                    $15.94$ &                    $15.43$ &           $\mathbf{15.18}$ &                    $15.21$ \\
\hline
\quad$\mathfrak{D}\!\left(0.7,0.6\right)$ &
         $18.63$ ($19.12$) &          $17.44$ ($17.78$) & $\mathbf{16.72}$ ($17.21$) &          $16.89$ ($16.98$) &          $17.39$ ($17.71$) &          $17.24$ ($17.41$) \\
\hline
\quad\quad$+$Ensemble                     &
                       $-$ &                    $16.37$ &                    $15.20$ &                    $15.11$ &                    $14.93$ &           $\mathbf{14.47}$ \\
\hline
\end{tabular}}
\begin{tabular}{|L{\colwidthB}|R{\colwidthC}||L{\colwidthB}|R{\colwidthC}||L{\colwidthB}|R{\colwidthC}|}
\hline
\cite{zhang2017interleaved} & $19.25$ &
\cite{huang2017snapshot} & $17.40$ &
\cite{han2017deep} & $17.01$ \\
\hline
\cite{zhang2017mixup} & $16.80$ &
\cite{gastaldi2017shake} & $15.85$ &
\cite{furlanello2018born} & $14.90$ \\
\hline
\end{tabular}
\caption[Caption for LOF]{Classification error rates ($\%$) by different models on CIFAR100. In each group, all networks have the same depth. Gen \#0 stands for the patriarch. We use a GitHub repository\footnotemark\ as our baseline (the numbers are from this website), and our re-implementation results are comparable to those reported originally ($22.80$ for DenseNet-$100$ and $17.17$ for DenseNet-$190$). Following conventions, we report accuracies from both the best epoch and the last epoch (numbers in parentheses), and all listed competitors reported the best epoch. {\bf We use the last epochs from each generation for ensemble.}}
\label{Tab:CIFAR100}
\end{table*}
\footnotetext{{\tt github.com/bearpaw/pytorch-classification}}

\noindent
$\bullet$\quad{\bf Setting and Baselines}

\noindent
We first evaluate our approach on the CIFAR100 dataset~\cite{krizhevsky2009learning}, which contains $60\rm{,}000$ tiny RGB images of a spatial resolution of $32\times32$. These images are split into a training set of $50\rm{,}000$ samples and a testing set of $10\rm{,}000$ samples. Both training and testing images are uniformly distributed over $100$ classes. We do not perform experiments on the CIFAR10 dataset because it does not contain fine-level classes, so that teacher-student optimization does not bring significant benefits (this was also observed in~\cite{furlanello2018born}).

We start with deep residual networks~\cite{he2016deep} with $20$, $56$ and $110$ layers, respectively. Let ${L}={6L'+2}$ layers where $L'$ is an integer. The first $3\times3$ convolutional layer is first performed on the $32\times32$ input image without changing its spatial resolution, then three stages followed, each of which has $L'$ residual blocks (two $3\times3$ convolutions plus one identity connection). Batch normalization~\cite{ioffe2015batch} and ReLU activation~\cite{nair2010rectified} are added after each convolution. The spatial resolution of the input remain unchanged in the three stages ($32\times32$, $16\times16$ and $8\times8$), and the number of channels are $16$, $32$ and $64$, respectively. Average pooling layers are inserted after the first two stages for down-sampling. The network ends with a fully-connected layer with $100$ outputs.

Our approach is also experimented on the densely connected convolutional networks (DenseNets)~\cite{huang2017densely} with $100$ and $190$ layers. The overall architecture is similar to the ResNets, but the building blocks are densely-connected, {\em i.e.}, each basic unit takes the input feature, convolves it twice, and concatenates it to the original feature. We use the $100$-layer and $190$-layer DenseNets with the standard numbers of base channels and growth rates.

We follow the conventions to train these networks from scratch. We use the standard Stochastic Gradient Descent (SGD) with a weight decay of $0.0001$ and a Nesterov momentum of $0.9$. In the ResNets, we train the network for $164$ epochs with mini-batch size of $128$. The base learning rate is $0.1$, and is divided by $10$ after $82$ and $123$ epochs. In the DenseNets, we train the network for $300$ epochs with a mini-batch size of $64$. The base learning rate is $0.1$, and is divided by $10$ after $150$ and $225$ epochs. In the training process, the standard data-augmentation is used, {\em i.e.}, each image is padded with a $4$-pixel margin on each of the four sides. In the enlarged $40\times40$ image, a subregion with $32\times32$ pixels is randomly cropped and flipped with a probability of $0.5$. No augmentation is used at the testing stage.

\vspace{0.2cm}
\noindent
$\bullet$\quad{\bf Results}

\noindent
We first evaluate the performance with respect to different hyper-parameters, namely, different parameterized processes $\mathfrak{D}\!\left(u\!\left(\eta\right),\lambda\right)$. We fix ${K}={5}$ and ${u\!\left(\eta\right)}={0.6}$, and diagnose the impact of $\lambda$ on deep residual networks~\cite{he2016deep} with different numbers of layers. We also evaluate the born-again networks~\cite{furlanello2018born} which corresponds to $\mathfrak{D}\!\left(1.0,0.5\right)$.

Results on deep ResNets are summarized in Figure~\ref{Fig:CIFAR100ResNets}. We can observe several important properties of our algorithm. First, a strict teacher ({\em i.e.}, the born-again network~\cite{furlanello2018born}, $\mathfrak{D}\!\left(1.0,0.6\right)$) is inferior to a tolerant teacher ({\em e.g.}, $\mathfrak{D}\!\left(0.6,0.6\right)$). Although the latter often starts with a lower accuracy of the patriarch model, it has the ability of gradually and persistently growing up and outperforming the baseline after $1$--$3$ generations. Meanwhile, recognition accuracy often saturates after a few generations, because eventually the teacher signal will converge to the points that are dominated by the primary classes ({\em i.e.}, ${f_{a_1}^{\left(m\right)}}\rightarrow{1}$), and the teacher will become strict again. However, the saturated accuracy is still much higher than the baseline, which demonstrates the reliability of our approach, {\em i.e.}, even if we cannot terminate at the best generation, we can still surpass the baseline.

In DenseNets with $100$ and $190$ layers, we report both single-model and model-ensemble results in Table~\ref{Tab:CIFAR100}. We evaluate $\mathfrak{D}\!\left(0.6,0.6\right)$ and $\mathfrak{D}\!\left(0.7,0.6\right)$, and observe the same phenomena as in ResNet experiments. In particular, in DenseNet-$100$, our single-model accuracy is $1\%$--$2\%$ higher, and our $5$-model ensemble accuracy is more than $5\%$ higher, even getting close to a single DenseNet-$190$ model. Considering DenseNet-$190$ requires around $30\times$ FLOPs of DenseNet-$100$, this is quite an efficient method to achieve high classification accuracy. In DenseNet-$190$, our results are competitive among the state-of-the-arts. Note that~\cite{zhang2017mixup} and~\cite{gastaldi2017shake} applied complicated data augmentation approaches to achieve high accuracy, but we found a different way, which is to improve the optimization algorithm.

\subsection{The ILSVRC2012 Dataset}
\label{Experiments:ILSVRC2012}

\renewcommand{\tabcolsep}{0.08cm}
\renewcommand{\colwidthA}{1.00cm}
\begin{table*}[!btp]
\centering
\begin{tabular}{|l||R{\colwidthA}|R{\colwidthA}|R{\colwidthA}|R{\colwidthA}|R{\colwidthA}|R{\colwidthA}|
R{\colwidthA}|R{\colwidthA}|R{\colwidthA}|R{\colwidthA}|R{\colwidthA}|R{\colwidthA}|}
\hline
{} &
\multicolumn{2}{c|}{Gen \#0} & \multicolumn{2}{c|}{Gen \#1} & \multicolumn{2}{c|}{Gen \#2} &
\multicolumn{2}{c|}{Gen \#3} & \multicolumn{2}{c|}{Gen \#4} & \multicolumn{2}{c|}{Gen \#5} \\
\hline\hline
Baseline                               &
$30.50$ & $11.07$ &     $-$ &     $-$ &     $-$ &     $-$ &     $-$ &     $-$ &     $-$ &     $-$ &              $-$ &              $-$ \\
\hline\hline
\:$\mathfrak{D}\!\left(0.6,0.6\right)$ &
$32.52$ & $11.23$ & $30.28$ & $10.23$ & $30.12$ & $10.15$ & $29.92$ & $10.25$ & $29.77$ & $10.19$ & $\mathbf{29.60}$ & $\mathbf{10.11}$ \\
\hline
\:\:$+$Ensemble                        &
    $-$ &     $-$ & $30.01$ & $ 9.98$ & $28.94$ & $ 9.53$ & $28.51$ & $ 9.36$ & $28.23$ & $ 9.28$ & $\mathbf{28.08}$ & $\mathbf{ 9.23}$ \\
\hline
\end{tabular}
\caption{Classification error rates (top-$1$ and top-$5$, $\%$) by different ResNet-$18$ models on ILSVRC2012.}
\label{Tab:ILSVRC2012}
\end{table*}

\noindent
$\bullet$\quad{\bf Setting and Baselines}

\noindent
With the knowledge and parameters learned from the CIFAR100 experiments, we now investigate the ILSVRC2012 dataset~\cite{russakovsky2015imagenet}, a popular subset of the ImageNet database~\cite{deng2009imagenet}. There are $1\rm{,}000$ classes in total. The training set and testing set contains $1.3\mathrm{M}$ and $50\mathrm{K}$ high-resolution images, with each class having approximately the same number of training images and exactly the same number of testing images.

We set the $18$-layer residual network~\cite{he2016deep} as our baseline. The $224\times224$ input image is passed through a $7\times7$ convolutional layer with a stride of $2$, and a $2\times2$ max-pooling layer. Then four stages followed, each with $2$ standard residual blocks (two $3\times3$ convolution layers plus an identity connection). The spatial resolution of these four stages are $56\times56$, $28\times28$, $14\times14$ and $7\times7$, and the number of channels are $64$, $128$, $256$ and $512$, respectively. Three max-pooling layers are inserted between these four stages. The network ends with a fully-connected layer with $1\rm{,}000$ outputs.

All networks are trained from scratch. We follow~\cite{hu2018squeeze} in configuring the following parameters. Standard Stochastic Gradient Descent (SGD) with a weight decay of $0.0001$ and a Nesterov momentum of $0.9$ is used. There are a total of $100$ epochs in the training process, and the mini-batch size is $1024$. The learning rate starts with $0.6$, and is divided by $10$ after $30$, $60$ and $90$ epochs. In the training process, we apply a series of data-augmentation techniques, including rescaling and cropping the image, randomly mirroring and rotating (slightly) the image, changing its aspect ratio and performing pixel jittering. In the testing stage, the standard single-center-crop is used on each image.

We inherit the best parameters learned from CIFAR100 experiments to ILSVRC2012 (the costly computation avoids us from tuning the hyper-parameters). We use $\mathfrak{D}\!\left(0.6,0.6\right)$, as the basic network ($18$ layers) is not very deep).

\vspace{0.2cm}
\noindent
$\bullet$\quad{\bf Results}

\noindent
Following CIFAR100 experiments, we set ${K}={5}$, ${u\left(\eta\right)}={0.6}$ and ${\lambda}={0.6}$. Results are summarized in Table~\ref{Tab:ILSVRC2012}. One can observe very similar results as in the previous experiments, {\em i.e.}, we start with a worse patriarch\footnote{It is interesting yet expected that the top-$1$ accuracy of the patriarch is $1.98\%$ lower than the baseline, but the top-$5$ accuracy is merely $0.16\%$ lower. This is because setting ${K}={5}$ hardly impacts top-$5$ classification.}, enjoy gradual and persistent improvement from generation to generation, and achieve saturation after several generations.

Limited by computational resources, we did not evaluate deeper networks. Although the performance of ResNet-$18$ is not directly comparable to deeper networks ({\em e.g.}, ResNets with $50$, $101$, $152$ layers or DenseNet with $264$ layers), our approach achieves a larger accuracy gain ($0.90\%$ top-$1$ and $0.96\%$ top-$5$) than other two light-weighted modules on ResNet-$18$, namely Squeeze-and-Excitation (SE)~\cite{hu2018squeeze} ($0.72\%$ top-$1$ and $0.80\%$ top-$5$) and Second-Order Response Transform (SORT)~\cite{wang2017sort} ($0.55\%$ top-$1$ and $0.27\%$ top-$5$). Compared to them, our approach does not require any additional computation at the testing stage, although the training stage is longer.

\vspace{0.2cm}
\noindent
$\bullet$\quad{\bf Transfer Experiments}

\renewcommand{\colwidthA}{1.30cm}
\begin{table}[!btp]
\centering
\begin{tabular}{|l||R{\colwidthA}|R{\colwidthA}||R{\colwidthA}|R{\colwidthA}|}
\hline
{}       & \multicolumn{2}{c||}{ILSVRC2012}    & \multicolumn{2}{c|}{Transfer}                          \\ 
\cline{2-5}
{}       & Top-$1$          & Top-$5$          & {\bf C256}       & {\bf I67}        \\
\hline\hline
Baseline & $69.50$          & $88.93$          & $75.94$          & $62.92$          \\
\hline\hline
Gen \#0  & $67.48$          & $88.77$          & $75.15$          & $62.97$          \\
\hline
Gen \#1  & $69.72$          & $89.77$          & $76.83$          & $62.75$          \\
\hline
Gen \#2  & $69.88$          & $89.85$          & $77.28$          & $\mathbf{65.72}$ \\
\hline
Gen \#3  & $70.08$          & $89.75$          & $77.65$          & $63.75$          \\
\hline
Gen \#4  & $70.23$          & $89.81$          & $77.00$          & $63.81$          \\
\hline
Gen \#5  & $\mathbf{70.40}$ & $\mathbf{89.89}$ & $\mathbf{77.83}$ & $63.94$          \\
\hline
\end{tabular}
\caption{Classification accuracies ($\%$) by different models on Caltech256 ({\bf C256}) and MIT Indoor-67 ({\bf I67} datasets. Top-$1$ and top-$5$ classification accuracies ($\%$) on ILSVRC2012 are also listed for reference.}
\label{Tab:TransferClassification}
\end{table}

\noindent
Finally, we transfer the trained models to feature extraction. We consider two popular datasets, namely, Caltech256~\cite{griffin2007caltech} and MIT Indoor-67~\cite{quattoni2009recognizing} for generic object classification and indoor scene classification, respectively. We follow the conventional settings, {\em i.e.}, extracting neural responses from the penultimate layer of the $18$-layer ResNet, and training a linear SVM to classify these $512$-dimensional vectors. We use the same training/testing split for all models, with $60$ and $80$ training images per class for Caltech256 and MIT Indoor-67, respectively.

Results are summarized in Table~\ref{Tab:TransferClassification}. We can find that, besides achieving better recognition accuracy on ILSVRC2012, the networks obtained by our approach indeed produce higher-quality transferrable features. These transfer experiments also verify that setting ${K}={5}$ generalizes well to other recognition tasks, {\em i.e.}, the intrinsic benefit comes from the secondary information, and our approach does not rely on a specific $K$. We believe these models also perform better in other transfer learning scenarios, {\em e.g.}, being fine-tuned on the PascalVOC or MS-COCO datasets.

\section{Conclusions}
\label{Conclusions}

In this work, we study the problem of optimizing deep networks in generations. This problem is meaningful, because it allows us to explore network optimization in depth; it is also useful, as we obtain better networks with the same architecture -- although training time becomes longer, testing time (including in transfer experiments) remains unchanged.

Based on the existing works, we provide a new viewpoint that teacher models should preserve secondary information, so that the students become stronger. We quantify these information, and empirically verify its impact in image classification. We train some standard networks on image classification datasets, and then transfer them to other recognition tasks. Our approach surpasses the single-generation and multi-generation baselines in every single case.

This research votes for the viewpoint that network optimization is far from perfect at the current status. In the future, we will investigate a more generalized model, including using a variable function at each generation and allowing $K$ to vary from case to case. In addition, we will consider a {\em temperature} term in Eqn~\eqref{Eqn:StudentLoss} to adjust the KL-divergence. Both are expected to achieve better optimization results.

{\small
\bibliographystyle{ieee}
\bibliography{egbib}
}

\end{document}